\documentclass[conference]{IEEEtran}
\IEEEoverridecommandlockouts

\usepackage{cite}
\usepackage{amsmath,amssymb,amsfonts}
\usepackage{algorithmic}
\usepackage{graphicx}
\usepackage{textcomp}
\usepackage{xcolor}
\usepackage{booktabs}
\usepackage{hyperref}
\usepackage{float}
\usepackage{subcaption}
\def\BibTeX{{\rm B\kern-.05em{\sc i\kern-.025em b}\kern-.08em
    T\kern-.1667em\lower.7ex\hbox{E}\kern-.125emX}}
\begin{document}

\title{Thought-For-Food: Reasoning Chain Induced Food Visual Question Answering\\
}

\author{\IEEEauthorblockN{Riddhi Jain}
\IEEEauthorblockA{
\textit{TCS-Research}\\
Pune, India  \\
jain.riddhi1@tcs.com}
\and
\IEEEauthorblockN{Manasi Patwardhan}
\IEEEauthorblockA{
\textit{TCS-Research}\\
Pune, India  \\
manasi.patwardhan@tcs.com}
\and
\IEEEauthorblockN{Parijat Deshpande}
\IEEEauthorblockA{
\textit{TCS-Research}\\
Pune, India  \\
parijat.deshpande@tcs.com}
\and
\IEEEauthorblockN{Venkataramana Runkana}
\IEEEauthorblockA{
\textit{TCS-Research}\\
Pune, India  \\
venkat.runkana@tcs.com}

}


\maketitle

\begin{abstract}

The immense diversity in the culture and culinary of Indian cuisines calls attention to the major shortcoming of the existing Visual Question Answering(VQA) systems which are inclined towards the foods from western region
Recent attempt towards building a VQA dataset for Indian food 
is a step towards addressing this challenge.
However, their approach towards VQA  follows a two-step process in which the answer is generated  first, followed by the explanation of the expected answer.
In this work, we claim that food VQA requires to follow a multi-step reasoning process  to arrive at an accurate answer, especially in the context of India food, which
involves understanding complex culinary context and identifying relationships between various food items. With this hypothesis we create reasoning chains upon the QA with minimal human intervention. 
We  fine-tune smaller LLMs and VLMs with auto-validated reasoning chains and further train them using reinforcement learning with larger data.  
With augmentation of reasoning chains, we observed accuracy improvement of an average 10 percentage points on the baseline.
We provide detailed analysis in terms the effect of addition of reasoning chains for the Indian Food VQA task.
\end{abstract}


\section{Introduction}
\label{sec:intro}

One of the most important part of culture and social aspects in everyday life is food.   In a country like India, food highlights immense diversity based on geography, religion, and traditions of different regions. A single meal
contain items  which differ in preparation, presentation and flavor. This richness in the culinary and the culture, poses unique set of challenges for AI systems that target the understanding of content related to Indian food.

A powerful framework that has emerged to connect visual and language reasoning is Visual Question Answering(VQA)\cite{antol2015vqa}. It has multiple applications in different areas of security\cite{security}, medical assistance\cite{medical} and also, culinary education\cite{agarwal2024indifoodvqa}. 
In food domain, VQA can act as an cooking assistant, nutritional analysis based on visual cues from the images directly.
However, the existing Food VQA mainly focuses on the cuisines from the West\cite{winata2024worldcuisines}, which restricts the AI systems to assist on Indian Food. This leaves a significant gap in the field with the non-western cuisines. Performance improvements of Indian food VQA is beneficial for multiple downstream usecases, such as Indian food recognition and analysis,  nutritional tracking, meal recommendation,  cultural education, etc.

\begin{figure*}
    
\centering
\includegraphics[width=\textwidth]{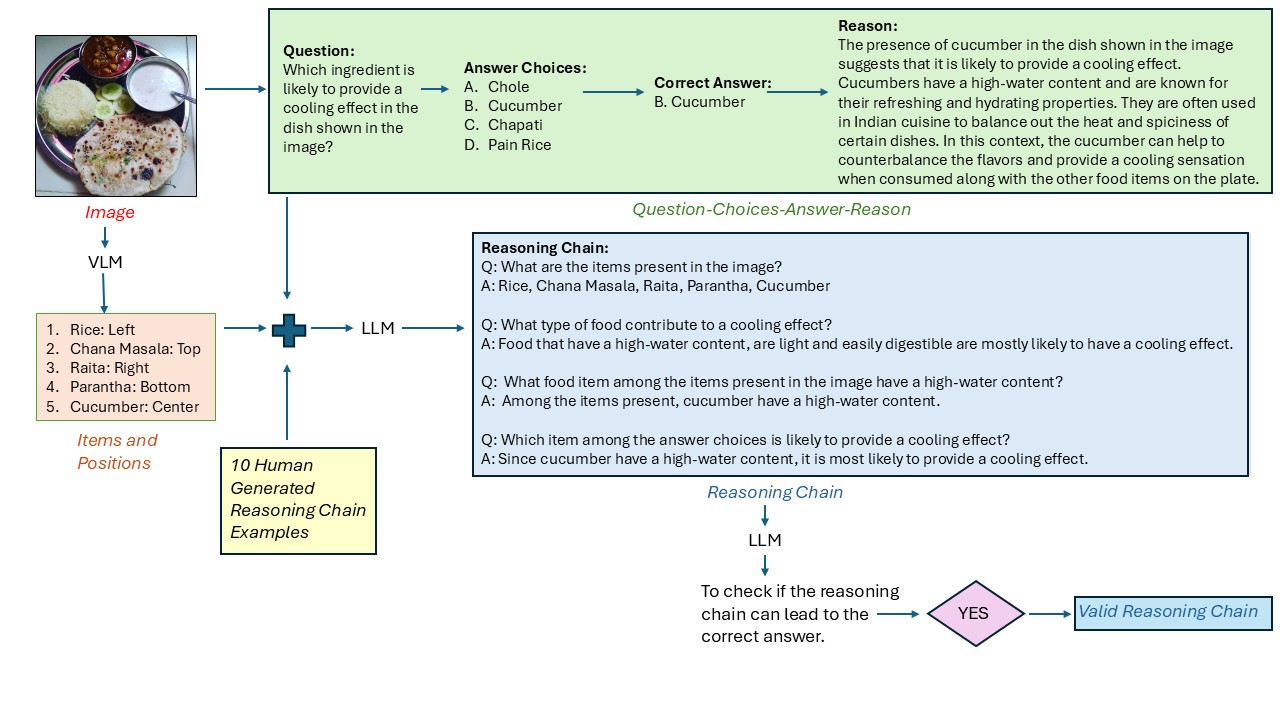}  
\caption{Valid Reasoning Chain Generation Method}
\label{ReasoningChain_l}
\end{figure*}

Recent progress has been made with the IndiFoodVQA dataset\cite{agarwal2024indifoodvqa}. It introduces VQA tasks for Indian food images, geared towards establishing a VQA dataset for a variety of question types specifically for Indian food. It evaluates different multi-modal language models under zero-shot and fine-tuned settings which sets an excellent foundation for the field. However, as discussed, the Indian food presents complexities beyond simple recognition. Images often contain multiple food items, and the question related to their placement, different nutritional and cultural information or cooking techniques requires image understanding as well as domain knowledge. To answer questions  the models might need to produce well-established reasoning chains, which ultimately lead to the final answer. Figure \ref{ReasoningChain_l}  demonstrates an example of such reasoning oriented question on an image of an Indian meal and the corresponding reasoning chain. 


IndiFoodVQA\cite{agarwal2024indifoodvqa} provides a `reason' explaining the right choice of the answer for a given question. The defined approach, to train models, utilizes this annotation as ground-truth `explainability' outcome after generating the answer.  On the other hand, we synthesize step-wise reasoning chains for each question, which lead to the answer with little assistance from humans and utilize these chains to train reasoning models.  
The IndiFoodVQA dataset incorporates hierarchical links between food products, their ingredients, and contextual factors into a related knowledge network in addition to direct visual reasoning. We use this to enhance the reasoning chain generation process, making sure that the models can include semantic associations present in the meal in addition to surface-level item and position detection. For example, as shown in Figure \ref{ReasoningChain_l}, we extract the annotation information, and pass it along with the few-shot reasoning chains, to produce the chains for the complete dataset. 

Using the IndiFoodVQA dataset enhanced with synthesized reasoning chains, we investigate the application of smaller LLMs and VLMs for the VQA task. We devise a strategy to validate synthesized reasoning chains, and use them to fine-tune models (Section \ref{def}). We further  train the models using more training samples with reinforcement learning (RL) to generate reasoning chains leading to an answer, rewarding the model based on correctness of the answer (Section \ref{rl}).

Our strategy consistently outperforms the IndiFoodVQA baseline, demonstrating the value of reasoning-based approaches for food VQA in culturally varied contexts (Section \ref{r&d}).

Our main contributions are listed as follows:
\begin{enumerate}
    \item We introduce a novel method to automatically generate reasoning chains tailored for Indian food visual question answering, addressing the cultural and culinary complexity of the domain as shown in Figure \ref{ReasoningChain_l}.
    
    \item We synthesize and validate reasoning chains  using multimodal language model with minimal human intervention, enabling structured multi-step reasoning over both visual and textual inputs, forming our synthetic supervised fine-tuning (SFT) dataset.
    
    \item Our method of SFT training followed by RL training outperforms previous baselines by around 10 percentage points and shows the efficacy of reasoning-driven approaches, achieving state-of-the-art scores on the IndiFoodVQA benchmark with an accuracy of 71.12\% by the best performing model.
\end{enumerate}

\section{Related Work}


\subsection{Food VQA}
Classification, retrieval, and recipe comprehension have been the primary focus of food-related AI research \cite{bossard2014food, salvador2019inverse}. With standards like WorldCuisines VQA \cite{winata2024worldcuisines} seeking to increase cultural coverage, food VQA has just lately attracted notice. The first systematic attempt to simulate Indian cuisine with VQA is IndiFoodVQA \cite{agarwal2024indifoodvqa}, which uses a two-step framework of response prediction followed by explanation generation. Building on IndiFoodVQA, we also make an effort in working towards building a systematic VQA for Indian food with an addition of reasoning chains generation. 


\subsection{Reasoning Chains}
Through chain-of-thought prompting, explicit reasoning has been shown to improve LLM performance on multi-step tasks \cite{tan2023boosting}. Multimodal reasoning extensions that try to ground reasoning in pictures include the visual chain-of-thought \cite{shao2024visual}. The ability of big language models to execute complicated reasoning is greatly enhanced by producing a chain of thought through a sequence of intermediate reasoning steps, as demonstrated in \cite{NEURIPS2022_9d560961}. \cite{occhipinti1994reasoning} work on strategies that are used to reason about food and contamination. \cite{tanabe2025reasoning} develops reasoning-driven food energy estimation using multi-modal large language models. We use this reasoning chain method, to solve the complexities in varied Indian Food using different multi-modal models with minimum human intervention.  

\subsection{Reinforcement Learning for LLMs and VLMs}
Reinforcement learning has emerged as a key component for matching language models to task-specific accuracy and human preferences \cite{ouyang2022training, zhong2024dpo}. RL has been investigated to impose visual grounding and lessen hallucinations in multimodal situations \cite{zhao2025iadgpt}. Its application to Food VQA has not yet been investigated, though, which is exactly what we have done in out work. We build upon our fine-tuned model with RL using different paradigms along with the reasoning chains.


\section{Dataset}
\label{dataset}
The IndiFoodVQA dataset\footnote{\hyperlink{https://github.com/SLSravanthi/IndifoodVQA}{https://github.com/SLSravanthi/IndifoodVQA}}, a benchmark curated for visual question answering in the context of Indian cuisine \cite{agarwal2024indifoodvqa}, serves as the foundation for our work. The dataset consists of total 16.7k samples (The dataset has
been split into the train, validation, and test sets in
a ratio of 70 : 10 : 20, thus consisting of 
709, 1661, and 3346 questions), where each sample consists of  an image of food, a question, a response, and a reason. The dataset is appropriate for assessing both recognition and reasoning because the questions cover a wide range of topics, from dish identity and ingredient inquiries to cooking specifics and contextual elements. The questions are divided into 12 different question types, which all fall under the domain of computational gastronomy, like ingredients, cooking technique, nutritional information, ingredient substitutions, etc. The complete list is mentioned in the Table \ref{tab:qwen2p5_qtype}.


\section{Problem Definition}
\label{task}

\subsection{IndiFoodVQA}\label{org_def}
A sample in  IndiFoodVQA \cite{agarwal2024indifoodvqa} consists of: 
\begin{enumerate}
\item an image ($\mathbf{I}$) of Indian food dish,  
\item  a question ($\mathbf{Q}$), 
\item  a set of four answer choices ($\mathbf{A} = \{a_1, a_2, a_3, a_4\}$) from which the correct answer must be selected, 
\item  external knowledge ($\mathbf{K}$) which are an optional set of triples which are relevant to the image and question type, extracted from the IndiFoodKG \cite{}. These triples are in a "subject; relation; object" format. The knowledge can be provided in two different ways: No external knowledge ($\mathbf{K} = \emptyset$) or 1-hop knowledge triples ($\mathbf{K}_{\text{1-hop}}$), where the relation directly connects two entities by "relation", 
\item correct Answer ($\mathbf{a_c}$) and 
\item  reason ($\mathbf{R}$), which is an explanation justifying the choice of $\mathbf{a_c}$.
\end{enumerate}
With $\mathbf{I}$, $\mathbf{Q}$, $\mathbf{A}$  and optional $\mathbf{K}$ as  inputs, the task is  to predict $\mathbf{a_c}$ followed by $\mathbf{R}$.

\subsection{Extended  Definition}
\label{def}
We augment an original sample of IndiFoodVQA with a reasoning chain $\mathbf{COT}$. We manually construct reasoning chains for ten exemplar, covering all the question-types (Table \ref{tab:qwen2p5_qtype}) to serve as a guide for generation of synthetic reasoning chains for other samples in the dataset.  The samples given for different question types are mentioned in Tables \ref{ex1} and \ref{ex2}.


\begin{table*}[ht]
\centering
\caption{Human Generated Reasoning Chains\\
Images are present in Figure \ref{images}}
\label{ex1}

\begin{tabular}{|l|l|l|l|}
\hline
& \textbf{Question and Answer Choice} & \textbf{Correct Answer and Reason} & \textbf{Reasoning Chains}\\
\hline
     \ref{1}& What is the cultural significance of the & C. It is a special dish enjoyed during Navratri festivals.
        & Q: What kind of meal does the combination\\

      &dish consisting of plain rice, chole,  & Reason: The dish consisting of plain rice, chole, curd, &  of rice, chole, curd, cucumber,\\

       &curd, cucumber, and chapati in Indian  &  cucumber, and chapati holds cultural significance during  &  and chapati provide? A: This combination provides\\

       &festivals? A. It is a popular dish served   & Navratri is a nine-night Hindu festival dedicated to the & a balanced and nutritious vegetarian meal. \\

     &during Diwali celebrations. B. It is a    & worship of the goddess Durga. The dish with its &  Q: Which festival among the listed ones, have a \\

     &traditional dish prepared during Holi     &  combination of rice, chole, curd, cucumber, and chapati. & significance of this kind of meal? A: Navratri is often \\

    &festivities. C. It is a special dish enjoyed     &inclusion of cucumber, which is hydrating and provides & accompanied with fast, which make this kind  \\

    &during Navratri festivals. D. It is a    &dietary fiber, is particularly beneficial during fasting &of meal suitable to be consumed. Q: Why is this meal  \\

    &customary dish served during & periods. The dish represents celebration of seasonal .&   suitable for Navratri? A: Items like cucumber,   \\

    & Eid al-Fitr celebrations. &  produce and cultural traditions associated with Navratri&provides hydration and dietary fibre, which is \\

    & & festivities.& beneficial during fasting periods.\\
     \hline

    \ref{2}&Which chutney is made with turmeric  & C. Tomato, onion, and raw mango chutney From the  & Q: What is turmeric, and how is it commonly\\

    &as an ingredient? A. Coconut chutney &  image, we can see that there are different types of&  used in Indian cuisine? A: Turmeric is a widely\\

   & B. Green chutney C. Tomato, onion,&  chutneys arranged around the uttapam. One of them&  used spice in Indian cooking, for flavor and color.\\

    &and raw mango chutney D. Healthy &  is made with turmeric as an ingredient. Among the& Q: What is the color of Turmeric?\\

    &green tomato chutney &  given options, only chutney that contains turmeric is &A: Turmeric is known for its bright yellow color. \\

    &&  Tomato, onion, and raw mango chutney. Turmeric is  & Q: Which of the given chutneys exhibit this color?\\

    && known for its health benefits and is commonly used  & A: Tomato, onion, and raw mango chutney\\

    && in Indian cuisine. Therefore, the correct answer is C.& is typically yellow-orange color.\\

    &&  Tomato, onion, and raw mango chutney. &\\

    \hline

    \ref{3}&How does cooking technique impact & A. The cooking technique enhances vibrant green color& Q: What gives it palak paneer characteristic green\\

    &the color of the palakpaneer?A. The &of the palakpaneer. From the image, the palakpaneer has &color? A: The spinach imparts a distinctive, vibrant  \\

    &cooking technique enhances vibrant &   a vibrant green color, which is enhanced by the& green color to the dish.Q: What technique is  \\

    &green color of the palakpaneer. B. &  cooking technique. Boiling the spinach (palak)   &applied on spinach?  A: Spinach is boiled for  \\

    &cooking technique turns palakpaneer& helps to retain its natural green color, & cooking palak paneer. Q:  How does boiling spinach\\

    &into a pale yellow color. C.Cooking &resulting in a visually appealing dish.  &affect its color? A: Boiling spinach helps retain\\

   & technique gives reddish hue to palak& &  and intensify its natural green color,\\

  &  paneer. D. The cooking technique does && making the cooked spinach\\

  & not affect the color of the palakpaneer.&& appear more vibrant.\\
    \hline
    \ref{4}&How does the presence of green chillies& C. They add a spicy and fiery flavor to the dish. From &Q: What is the typical taste of green chillies?\\
    
    &contribute to the taste and flavor profile&image, it can be observed that the green chillies are &A: Green chillies are known for their spicy and fiery\\

    &of the dish? A. They add a mild and &placed alongside the other ingredients on the plate. &  flavor due to the presence of capsaicin,a compound \\

    &cooling flavor to the dish. B. They &The bright green color and slender shape of the chillies &  that gives heat to peppers.Q: What does the placement \\

    &add a tangy and citrusy flavor to the&indicate that they are fresh and potentially spicy. Green& of green chillies alongside other ingredients suggest\\

    & dish.C. They add a  spicy and fiery flavor& chillies are known for their heat and fiery flavor, which&  about how they’re meant to be consumed?A: Their \\

     &to dish.D. They add  bitter and earthy& adds a spicy kick to the dish. This flavor profile & placement near the main ingredients suggests they\\

     & flavor to dish.&enhances the overall taste and adds a level of heat & are meant  to be eaten with the dish  to influence its \\

     & &to the dish.& flavor. Q:How does green chillies affect the dish's \\

    &  && taste and flavor profile? A: Green chillies, being fresh\\

    &  &&  and spicy, add a fiery kick to the dish, enhancing its\\

    &  &&  overall flavor profile with heat and intensity.\\

    \hline

  \ref{5}&  What is nutritional benefit of including & A. Bitter gourd is fiber rich and helps digestion.&Q: What is key nutrition component of bitter gourd?\\

   & bitter gourd in this meal? A. Bitter gourd &  Bitter gourd is known for its high fiber content, &A:  Bitter gourd is high in dietary fiber, which is one \\

    &is high in fiber and helps in digestion. &which aids in digestion and promotes a healthy &of its most prominent nutritional qualities.Q: How \\

    &B. Bitter gourd is rich in protein and &digestive system. It helps in regulating blood &does fiber benefit the body? A: Dietary fiber aids in \\

   & helps muscle  growth. C. Bitter gourd is & sugar levels and is beneficial for weight & digestion by promoting regular bowel movements  \\

    &  good source of vitamin C and boosts &management. Other answer choices are not  & and supporting a healthy digestive tract.\\

    &immunity.D. Bitter gourd  is high in &accurate as bitter gourd is not particularly &\\

    &potassium and helps in regulating &high in protein, vitamin C, or potassium.&\\

    &blood pressure. &&\\
    \hline

\end{tabular}
    
\end{table*}

We refer to the expert provided `reason' ( $\mathbf{R}$) annotation for the exemplars to manually define the reasoning chains.  Every chain illustrates the systematic breakdown of the question and corresponding responses leading to the right answer.  

We want to utilize a reasoning model to synthesize reasoning chains for all the sample in the dataset with these manually annotated exemplars serving as the few-shots. We use a VLM (Qwen2-VL-7B-Instruct) to extract a  list of the food items and their locations from  the food image in a step-wise fashion (first extract the food items and then the approximate spacial locations).  A structured food item-position ($\mathbf{\{F-P\}}$) map is created as illustrated in Figure \ref{ReasoningChain_l}. For each sample in the original dataset, we provide this map along with the few-shot exemplars,  the original question–answer alternatives with the right answer and the explanation to a reasoning model  (DeepSeek-R1) and prompts (System Prompt: \textit{You are a reasoning model expert in forming structured reasoning to achieve the final result} and User Prompt is elaborated in Table \ref{prompts}) it to generate a reasoning chain that links textual and visual data to support the expected answer.

With the augmentation of the reasoning chains the newly defined task is:  with $\mathbf{I}$  or $\mathbf{\{F-P\}}$, $\mathbf{Q}$, $\mathbf{A}$  and optional $\mathbf{K}$ as  inputs,  generate $\mathbf{COT}$ followed by $\mathbf{a_c}$. We utilize $\mathbf{I}$  as an input when we use Visual Language Models (VLMs). Whereas we utilize the food item-position map  $\mathbf{\{F-P\}}$ extracted from the image $\mathbf{I}$ as an input, when we use a Large Language Models (LLMs) with reasoning capabilities, that can not accept image modality inputs. Given the set of inputs, prediction of the correct answer $\mathbf{a_c}$ is consistent across the original task definition explained in Section \ref{org_def} and the extended task definition. The approaches are evaluated on this answer prediction task.








\begin{table*}[tbp]
\caption{Human Generated Reasoning Chains\\
Images are present in Figure \ref{images}}
\label{ex2}
\begin{tabular}{|l|l|l|l|}
\hline
 &\textbf{Question and Answer Choice} & \textbf{Correct Answer and Reason} & \textbf{Reasoning Chains}\\
\hline

   \ref{6} &What can be inferred about seasonality&A. Desserts are likely to be consumed during winter& Q: Which ingredient from the options, might be present  \\

    &and locality of the Indian desserts in & months due to presence of garam masala, which is&in the food?A: The food might have garam masala in it.\\

    &image based on their appearance and & commonly used in warming dishes. Based on the &Q: What does garam masala contain?A:  It typically \\

   & ingredients? A. Desserts are likely to&appearance and ingredients of the Indian&contains warming spices like cinnamon, cloves, and \\

   & be consumed during winter due to& desserts in the image, the presence of garam masala &  black pepper.Q:  What kind of effect do the spices  \\

    &presence of garam masala, which is & these desserts are likely to be consumed&suggests in garam masala have on body? A: Spices in \\

    &commonly used in warm dishes.B. The& during winter. Garam masala is spice blend&garam masala are known to produce internal \\

    &desserts are to be consumed during & commonly used in Indian cuisine, especially during& warmth and are often used to generate \\

     &summer months due to use of fenugreek,& colder seasons, as it contains warming spices&heat in the body during colder seasons.\\

     &which is known for cooling properties.& like cinnamon, cloves, and black pepper. These &\\

     &C. Desserts are likely to be consumed&are known to generate heat in the body and provide.&\\

     &throughout year as they don't contain & a comforting and cozy flavor profile, making &\\

     &any season-specific ingredients. D. The&them ideal for winter consumption. Therefore, &\\

     &desserts to be consumed during autumn&option A is the correct answer.&\\

     &months due to presence of black pepper,&&\\
     
     &which is commonly used in fall-inspired &&\\

     &recipes.&&\\

\hline

    \ref{7}&What ingredient could be substituted in &C. Fruit custard. Based on the image, the white balls &Q: What are the white balls visible in the image?\\

    &white balls in the image to create a dish& are likely rasgulla. To create a dish with a different&A: The white balls are likely to be rasgulla an Indian \\

    &with a different texture and taste?& texture and taste, substituting the white balls&dessert made from chenna(paneer) and sugar syrup.\\

    &A. Saffron B. Cardamom C.  Fruit custard& with fruit custard would be a suitable option. Fruit& Q: What is the typical texture and taste of rasgulla?\\

    &D. Chicken razala&custard has a creamy texture and a fruity taste, which& A: Rasgullas have a soft and spongy texture with light, \\

    &&  would contrast with the soft and spongy texture of the& sweet taste.Q: Which item among the given options,\\

    && rasgulla. This substitution would add a different flavor& have profile different from rasgulla?A: Fruit\\

    && profile to the dish, enhancing the overall experience.& custard is a creamy dessert with a smooth texture and\\

    && a fruity, sweet flavor. & Q: Why would fruit custard be a suitable substitution\\

    &&&  among the given options?A: Fruit custard provides a\\

    &&&  distinctly different texture and flavor, offering a\\

    &&&  new experience compared to other options like saffron\\

    &&&  or cardamom, which only alter flavor, not texture,\\

    &&&  or chicken razala, which wouldn't fit a dessert context.\\
\hline

  \ref{8}  &How does placement of  medu vada on & B. The medu vada represents  arrival of monsoon; & Q: What are  main ingredients in medu vada? A: It is a \\

    &the plate reflect celebration of seasonal&  availability of lentils. The placement of the medu&deep-fried fritter, primarily made from urad dal(lentil). \\

    &produce?A.Meduvada signifies abundance& vada in  center of plate suggests its significance &Q:When do lentils become readily available in India?\\

    &of fresh vegetables during harvest season.& in the celebration of seasonal produce. During the&A: Lentils become widely available during the monsoon\\

    &B. The medu vada represents arrival of & monsoon season in India, lentils become readily& season in India, as it is favorable for their harvest.\\

    &monsoon and availability of lentils.C. & available, and medu vada, deep-fried lentil fritter,&Q: How is the medu vada typically positioned on the\\

    &Medu vada symbolizes summer and& is a popular dish prepared using these ingredients. & plate in given image, and what does this suggest?\\

    &bountiful harvest of grains.D. The medu& The positioning of the medu vada in the center&A:  In the image, the medu vada is placed prominently at\\

    &vada signifies  winter season and the & highlights its importance and connection to  arrival& the center of the plate, indicating its special importance\\
    
    &abundance of root vegetables.& of monsoon and the availability of lentils, making& in the dish.\\

    && it an integral part of the seasonal celebrations.&\\
\hline

   \ref{9} &What is possible fusion dish that can& C. Idli Vada Tacos. By combining idli and vada, &Q: What is a fusion dish? A: A fusion dish blends\\

    &be created by combining  idli and vada & a fusion dish like Idli Vada Tacos can be created.& elements from different cuisines.Q:What will be result \\

    &from  image? A. Idli Vada Burger B. Idli& The idli can be used as the taco shell, while the& of combinining idli and vada in this way?A: Combining \\

    &Vada Pizza C. Idli Vada Tacos D. Idli & vada can be placed inside as filling. This innovative& idli and vada introduces new textures and flavors,\\

    &Vada Sushi& fusion combines the soft and fluffy texture of idli & creating an innovative eating experience.Q: Among the \\

    &&with the crispy and savory vada, creating a unique&options, which fusion dish is possible? A: Taco is \\

    && culinary experience. The fusion of Indian and &possible, since idli can be used as a shell or boat,\\

    &&Mexican flavors adds a delightful twist to the& opened or molded to hold the vada, which acts as the \\

    && traditional idli and vada dishes.&filling inside.\\
\hline

   \ref{10} &Which food item in the image contains & D. Chutney. By examining image, it can be & Q: What are some common food allergens?A: The most\\

    &an allergenic ingredient that can be &determined that food item containing an allergenic  &  common food allergens are peanuts, dairy and wheat.\\

    &substituted with an alternative?&ingredient that can be substituted with an alternative &Q: Which food among the given options contains these\\

    & A. Vada B. Idli C. Dosa D. Chutney&is chutney. Chutneys include various ingredients, & common food allergens? A: Chutney usually contains\\

    & &and if those ingredients are allergenic, they can be&peanuts. Q: Can peanuts in chutney be substituted?\\ 

    & & replaced with suitable alternatives to accommodate&A: Yes, peanuts can be substituted by  sunflower seeds.\\

    & & dietary restrictions. Therefore, the correct answer&\\

    & &  is chutney. &\\

\hline

\end{tabular}
    
\end{table*}

\begin{figure*} [htbp]
    \centering
    \caption{Images for Human Generated Reasoning Chains in Tables \ref{ex1} and \ref{ex2}}
    \label{images}
    \begin{subfigure} [b]{0.18\textwidth}
        \includegraphics [width=\linewidth] {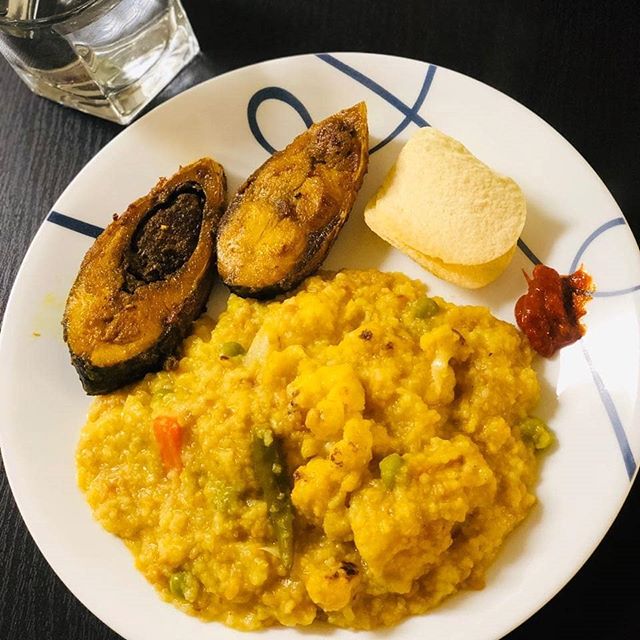}
        \caption{2a}
        \label{1}
    \end{subfigure}
    \begin{subfigure} [b]{0.18\textwidth}
        \includegraphics [width=\linewidth] {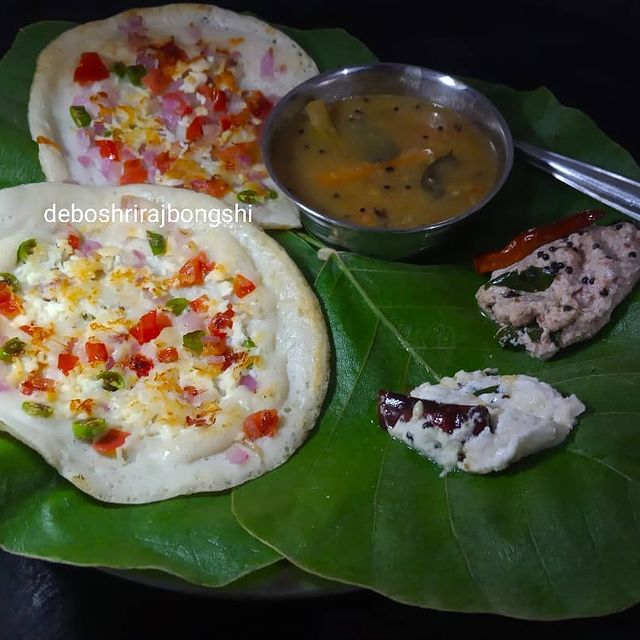}
        \caption{2b}
        \label{2}
    \end{subfigure}
    \begin{subfigure} [b]{0.18\textwidth}
        \includegraphics [width=\linewidth] {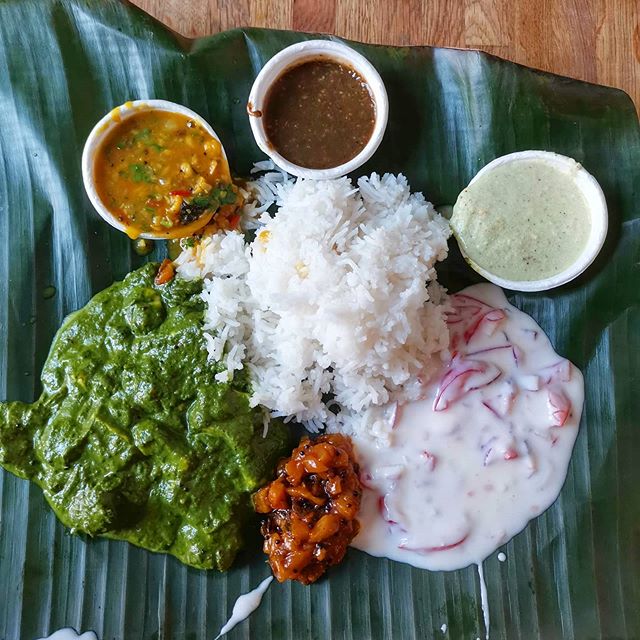}
        \caption{2c}
        \label{3}
    \end{subfigure}
    \begin{subfigure} [b]{0.18\textwidth}
        \includegraphics [width=\linewidth] {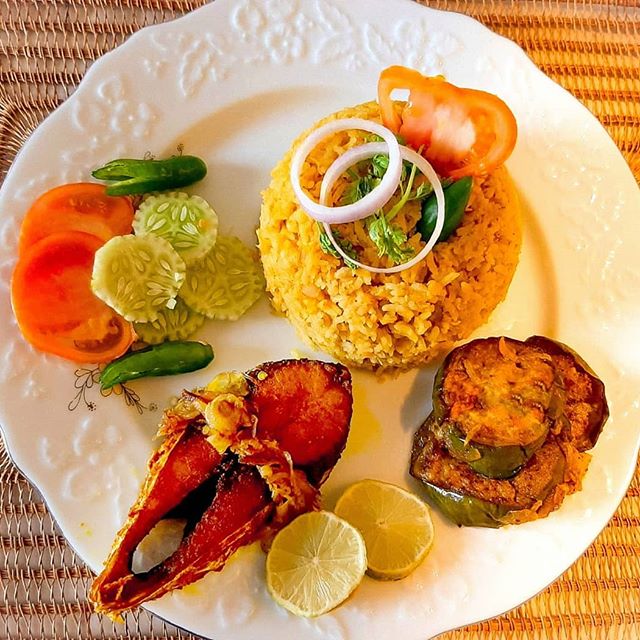}
        \caption{2d}
        \label{4}
    \end{subfigure}
    \begin{subfigure} [b]{0.18\textwidth}
        \includegraphics [width=\linewidth] {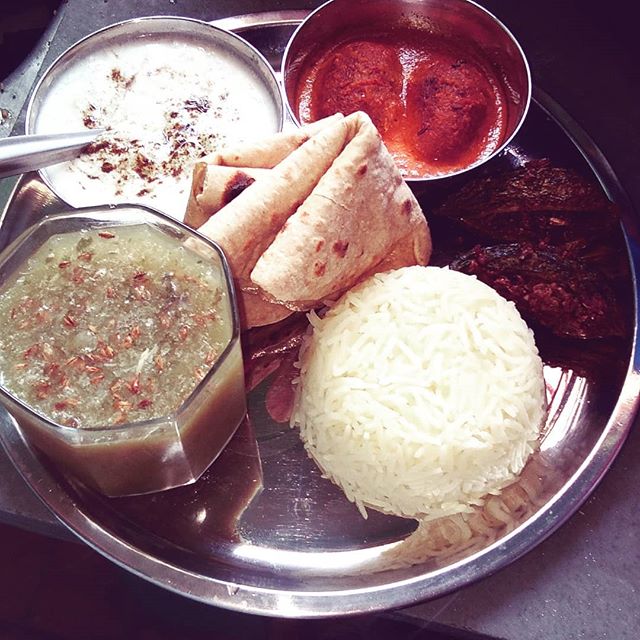}
        \caption{2e}
        \label{5}
    \end{subfigure}
\vspace{0.5cm}

    \begin{subfigure} [b]{0.18\textwidth}
        \includegraphics [width=\linewidth] {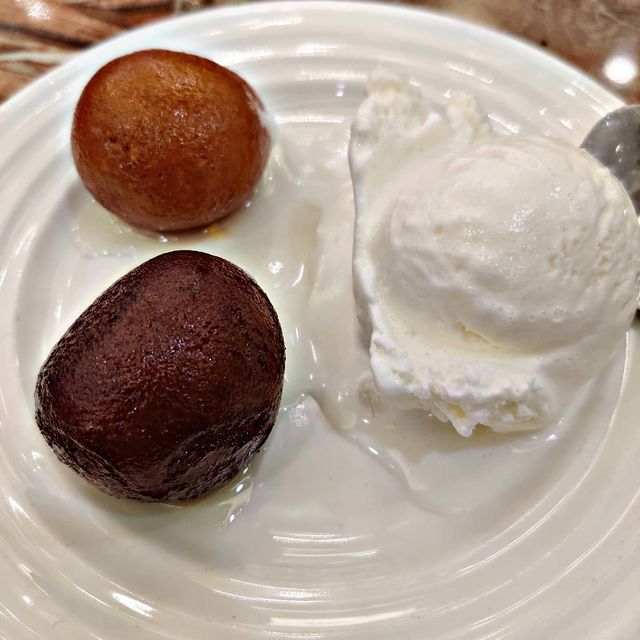}
        \caption{2f}
        \label{6}
    \end{subfigure}    
    \begin{subfigure} [b]{0.18\textwidth}
        \includegraphics [width=\linewidth] {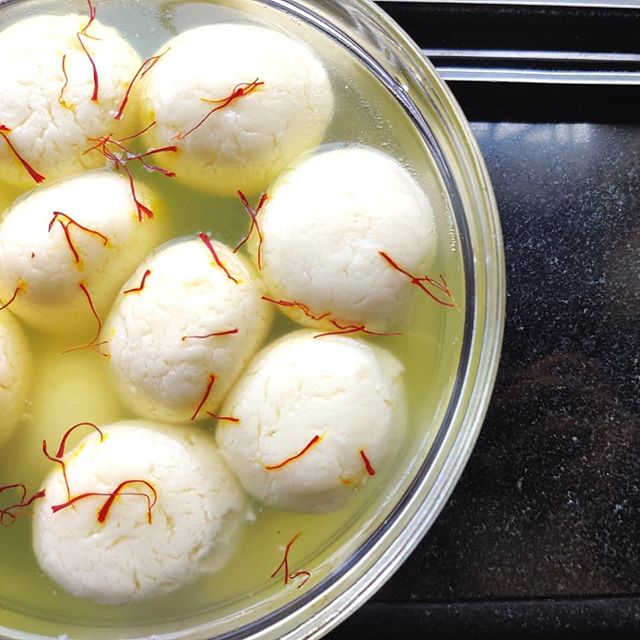}
        \caption{2g}
        \label{7}
    \end{subfigure}    
    \begin{subfigure} [b]{0.18\textwidth}
        \includegraphics [width=\linewidth] {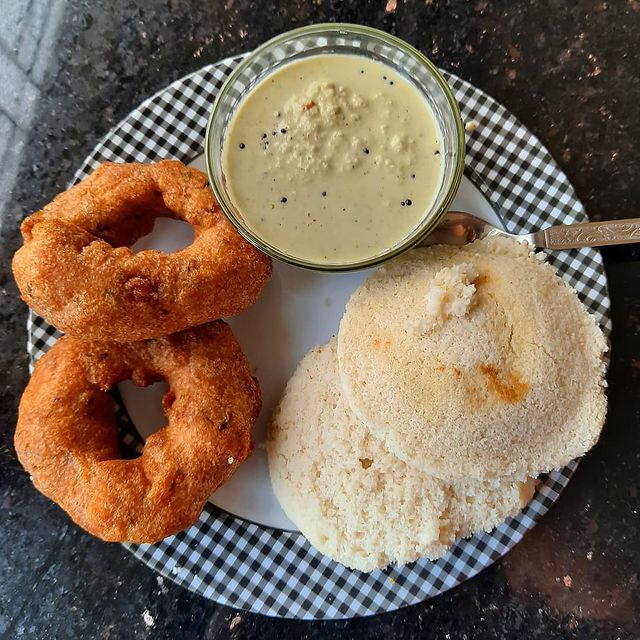}
        \caption{2h}
        \label{8}
    \end{subfigure}    
    \begin{subfigure} [b]{0.18\textwidth}
        \includegraphics [width=\linewidth] {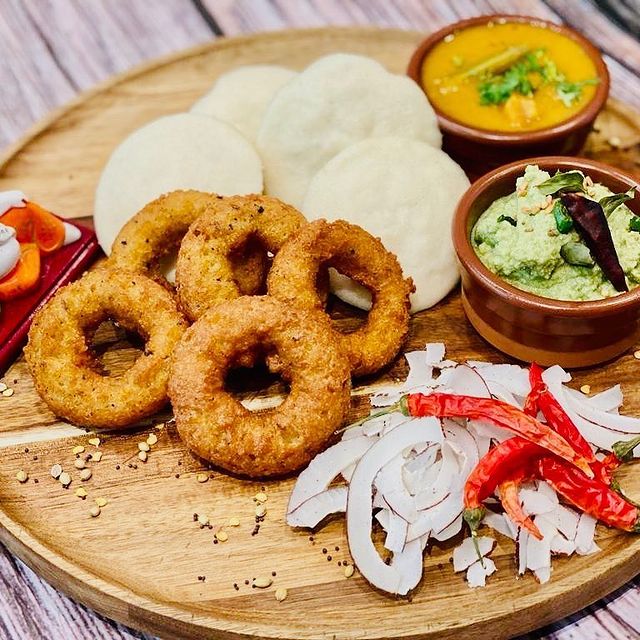}
        \caption{2i}
        \label{9}
    \end{subfigure}    
    \begin{subfigure} [b]{0.18\textwidth}
        \includegraphics [width=\linewidth] {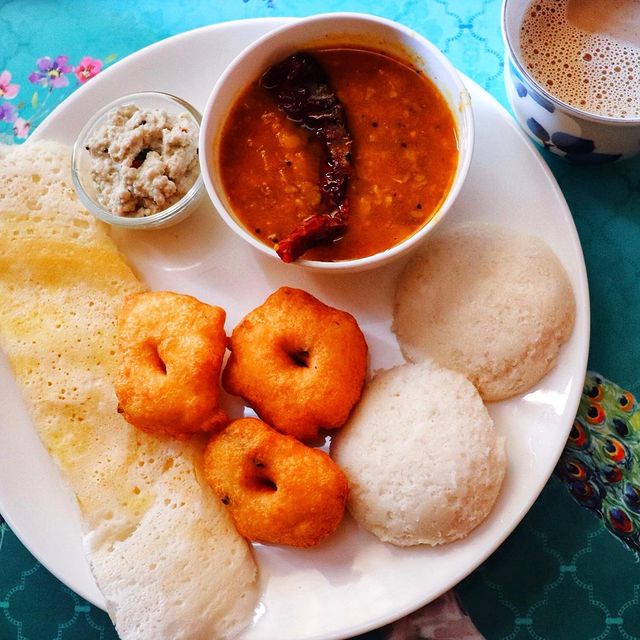}
        \caption{2j}
        \label{10}
    \end{subfigure}    

\end{figure*}

\section{Method}

Our method is divided into two stages: (i) Supervised Fine-Tuning (SFT) followed by (ii) Reinforcement Learning (RL). 

\subsection{Supervised Fine-Tuning}
To validate the synthesized reasoning chains (Section \ref{def}) of the samples of IndiFoodVQA dataset, we feed the food item-position pairs $\mathbf{\{F-P\}}$, the question $\mathbf{Q}$, answer choices $\mathbf{A}$  and  the synthesized stepwise reasoning chain $\mathbf{COT}$ as inputs to the reasoning model (DeepSeek-R1-0528-Qwen3-8B) and prompt it to choose the correct answer. The system prompt is: \textit{You are a Visual/Language assistant model, expert in Indian Food Analysis.} and detailed user prompt is illustrated in Table \ref{prompts}. The reasoning chain is are considered to be valid if the model can infer the correct answer $\mathbf{a_c}$. All the training samples of IndiFoodVQA dataset with valid synthesized reasoning chains are considered for supervised fine-tuning (SFT). This guarantees that the model learns to generated  logical, step-by-step explanations leading to correct answer rather than overfitting to incorrect reasoning chains, which do not lead to correct answers. Out of all the training samples, we observe  65.69\% samples had valid reasoning chains. This subset of the data is divided into two parts:   80\% for training and 20\% for validation.

We fine tune the LLMs and VLMs for the extended task definition (Section \ref{def}) with the validated samples of augmented IndiFoodVQA dataset. We perform parameter-efficient fine-tuning (PEFT), we use the LoRA (Low-Rank Adaptation)\cite{hu2022lora,zanella2024low} framework, which enables us to train large models within feasible computational budgets while maintaining their general capabilities. The prompt in SFT is same as the one used in zero-shot and is mentioned in the Table \ref{prompts}.
SFT trains the models to  predict answers while reinforcing logically coherent reasoning chains. Thus, we expect them to perform considerably better than the base model as well as the onse which are trained to directly predict the answer without the intermediate reasoning. 
    
    
    

\begin{table}[h]
    \centering
    \caption{List of Prompts}
    \label{prompts}
    \begin{tabular}{|l|l|}
    \hline
       Annotations  & Carefully examine the food plate\\
       & in this image and list the names\\
                    &  of all visible Indian food items \\
                    & or dishes. Do not describe or explain\\
                    &  the items; just provide a simple,\\
                    & comma-separated list of their names. Ensure \\ 
                    & each item is visually distinct and identifiable,\\ 
                    & noting that some items might be present on top \\
                    &of others. Also specify their relative positions.\\
    
        \hline
       Reasoning   & Given a multiple-choice question:\{question\}, \\
        Chain & and its answer: \{answer\},\\
         Generation& Decompose the reasoning into a series of \\
         & subquestions and subanswers. Each step\\
         &  should build on the previous step where applicable. \\
        \hline
       Answer &  Question: \{question\}; \\
       and Reason & Answer Choices: \{answer\_choices\};   \\
        Prompt        &  correct answer out of the four given choices.\\
                &  Provide a clear reason for the chosen answer.\\ 

        \hline
    \end{tabular}
    
\end{table}

    \begin{table*}[h]
    \centering
    \caption{Results: Accuracy (\%) of different models across training strategies. () - \% Improvement over Zero-shot \\
    \textbf{\underline{Bold and Underlined: Best Performance}};
    \textbf{Bold:  Best for the model}
    }
    \label{result}
    \small
    \begin{tabular}{|c|c|c|c|c|}
        \hline
        & \textbf{Llama-3.1-8B-Instruct} & \textbf{DeepSeek-R1-0528-Qwen3-8B} & \textbf{Qwen2.5-VL-3B-Instruct} & \textbf{Qwen2-VL-7B-Instruct} \\
          
   \hline
        Zero-Shot & 35.85  & 31.86 & 51.30 & 50.91\\
          \hline
        IndiFoodVQA\cite{agarwal2024indifoodvqa}:No KG& 52.74 (+16.89) & 46.82 (+14.96) & 62.21 (+10.91) & 58.83 (7.92)+\\
        \hline
        IndiFoodVQA\cite{agarwal2024indifoodvqa}: 1-hop & 50.21 (+14.36) & 46.00 (+14.14) & 59.11 (+7.81) & 55.39 (+4.48)\\
        \hline
        \hline
        SFT Training & 56.15 (+20.13)  & 51.90 (+20.04) & 65.37 (+14.07) & 65.00 (+14.09)\\
        \hline
        DPO RL Training& \textbf{69.00} (+33.15) & \textbf{62.14} (+30.28)  & \textbf{\underline{71.12}} (+19.82) & 68.50 (+17.59)\\
        \hline
        GRPO  RL Training& 66.02 (+30.02) & 59.64 (+27.78) & \textbf{70.43} (+19.13) & \textbf{69.92} (+19.01)\\
        \hline
        \hline
        SFT with KG & 55.91 (+20.06) & 49.00 (+17.14) & 61.33 (+10.03) & 64.53 (+13.62)\\
        \hline
        DPO with KG & 60.24 (+24.39) & 55.64 (+23.78) & 64.10 (+12.8) & 65.11 (+14.20)\\
        \hline
        GRPO with KG& 60.80 (+24.95)  & 54.16 (+22.3)  & 66.25 (+14.95) & 64.39 (+13.48)\\
        \hline
    \end{tabular}
    \end{table*}

   \subsection{Reinforcement Learning}
   \label{rl}
 Following SFT, we use reinforcement learning (RL) to further optimize both LLMs and VLMs with additional samples, including the ones for which we do not have the availability of valid reasoning chains. Thus we use the complete IndiFoodVQA  dataset for RL training, with the the original  80/20 split.  The prompt used for RL training is consistent with  the prompt of the SFT training. 

 We try two preference-optimization paradigms: \textbf{Direct Preference Optimization (DPO)}\cite{rafailov2023direct} and \textbf{Group Relative Preference Optimization (GRPO)}\cite{zhang2025r1}. 
 We establish a binary reward function for both DPO and GRPO that assesses if the reasoning chain results in the right response. Rewards are automatically calculated: $1$ is awarded for reasoning chains that end with the right solution, and $0$ is awarded for those that produce incorrect answers. 

 This  improves reasoning consistency and robustness by forcing the model to learn to generate reasoning chains that actually lead to right answers.

This reward function is used differently in DPO and GRPO, and is described below.  

 \subsubsection{Direct Preference Optimization (DPO)}  
In DPO, preference pairs that are obtained from reasoning chains are used to directly optimise the model. We pair reasoning chains that lead to the right answer with those that lead to the wrong answer for every question. While reducing the likelihood of less-favored chains, the training objective pushes the model to give the chosen (correct-answer) chain a larger likelihood.

In DPO, we train on pairs of reasoning chains $(\mathbf{C}^+, \mathbf{C}^-)$ for the same input $(\mathbf{I}, \mathbf{Q}, \mathbf{A}, \mathbf{K})$, where $\mathbf{C}^+$ leads to the correct answer $\mathbf{a_c}$ and $\mathbf{C}^-$ does not. The objective encourages the policy $\pi_\theta$ to assign higher probability to $\mathbf{C}^+$:

\begin{align}
\mathcal{L}_{\text{DPO}} = 
&- \mathbb{E}_{(\mathbf{I}, \mathbf{Q}, \mathbf{A}, \mathbf{K}, \mathbf{C}^+, \mathbf{C}^-)} \Bigg[ \\
& \quad \log \sigma \Big( \beta \cdot \big( \log \pi_\theta(\mathbf{C}^+\\
& \mid \mathbf{I}, \mathbf{Q}, \mathbf{A}, \mathbf{K}) 
 - \log \pi_\theta(\mathbf{C}^- \mid \mathbf{I}, \mathbf{Q}, \mathbf{A}, \mathbf{K}) \big) \Big) \Bigg] \nonumber
\end{align}

where $\pi_\theta$ is the model policy, $\sigma(\cdot)$ is the sigmoid function, and $\beta$ is a scaling parameter.

\subsubsection{Group Relative Preference Optimization (GRPO)}
In GRPO, the reinforcement learning is explicitly modelled during reasoning chain generation, extending the preference signal into the generation space. For a given input, the model samples several reasoning chains from comparable question types, and each chain is rewarded according to whether it results in the right answer in the end.

\[
r_i = \begin{cases}
1 & \text{if } f(\mathbf{C}_i, \mathbf{Q}, \mathbf{A}) = \mathbf{a_c}, \\
0 & \text{otherwise}.
\end{cases}
\]

We compute the relative advantage of $\mathbf{C}_i$ as:

\[
A_i = r_i - \frac{1}{m}\sum_{j=1}^m r_j.
\]

The GRPO objective is then defined as:

\[
\mathcal{L}_{\text{GRPO}} = - \mathbb{E}_{(\mathbf{I}, \mathbf{Q}, \mathbf{A}, \mathbf{K})} \left[ \sum_{i=1}^m A_i \cdot \log \pi_\theta(\mathbf{C}_i \mid \mathbf{I}, \mathbf{Q}, \mathbf{A}, \mathbf{K}) \right].
\]

This formulation reinforces reasoning chains with above-average rewards while penalizing those with below-average rewards, ensuring that the model improves its ability to generate valid reasoning over time.

\subsection{Augmentation of Domain Knowledge}
We extend the baseline work through the following augmentation steps. Firstly, we create structured annotations using VLM. Each image $\mathbf{I}$ is asked to produce: an item list $\mathbf{L}$ containing all food items in the image, and a position list $\mathbf{P}$ indicating their relative location.
\[
(\mathbf{I}, \mathbf{Q}, \mathbf{A}) \;\; \Rightarrow \;\; (\mathbf{I}, \mathbf{Q}, \mathbf{A}, \mathbf{L}, \mathbf{P}).
\]

Secondly, we augment the task by generating reasoning chain $\mathbf{C}$ that justifies the choice that modifies the learning objective to
\[
\mathbf{F}_{\text{ours}} : (\mathbf{I}, \mathbf{Q}, \mathbf{A}, \mathbf{L}, \mathbf{P}) \rightarrow (\mathbf{a_c}, \mathbf{C}),
\]
where $\mathbf{C}$ is the set of structured decompositions. We integrate reasoning chains into training and evaluation, using them as separate supervision.

\begin{table*}[h]
\centering
\caption{Question-type wise accuracy (\%) for Qwen2.5-VL across training strategies 
() - \% Improvement over IndiFoodVQA:No-KG
}
\label{tab:qwen2p5_qtype}
\begin{tabular}{|lcccc|}
\toprule
\textbf{Question Type} & \textbf{IndiFoodVQA} & \textbf{Fine-tuning} & \textbf{DPO} & \textbf{GRPO} \\
&\textbf{NO-KG}&&&\\
\midrule
Ingredients & 61.2 & 63.5 (+2.3) & 65.8 (+4.6) & 65.4 (+4.2) \\
\textbf{Cooking technique} & \textbf{60.5} & \textbf{68.0} (+7.5) & \textbf{68.9} (+8.4) & \textbf{68.6} (+8.1)\\
Cultural significance & 62.8 & 64.1 (+1.3) & 64.8 (+2.0) & 64.6 (+1.8)\\
Taste and flavor profile & 61.0 & 65.2 (+4.2) & 69.6 (+8.6) & 69.2 (+8.2) \\
Health and nutrition & 59.7 & 63.9 (+4.2) & 68.5 (+8.8) & 67.9 (+8.2)\\
Seasonality and locality & 60.2 & 64.0 (+3.8)& 69.4 (+9.2)& 69.0 (+8.8)\\
\textbf{Ingredient substitutions} & \textbf{58.5} &\textbf{64.3} (+5.8) & \textbf{71.5} (+13)& \textbf{72.1} (13.6)\\
Presentation and plating & 62.1 & 64.0 (+1.9) & 65.2 (+3.1)& 65.0 (+2.9)\\
\textbf{Fusion and innovation} & \textbf{57.9} & \textbf{65.0} (+7.1) & \textbf{71.8} (13.9)& \textbf{72.4} (14.5)\\
Cooking science & 58.8 & 63.7 (+4.9)& 70.6 (11.8)& 70.0 (11.2)\\
\textbf{Allergens and restrictions} & \textbf{57.5} & \textbf{62.9 }(+5.4)& \textbf{71.2} (13.7)& \textbf{71.9} (14.4)\\
Food pairings & 59.0 & 64.2 (+5.2)& 72.1 (13.1)& 71.7 (12.7)\\
\bottomrule
\end{tabular}
\end{table*}

Thirdly, we include reinforcement learning techniques like DPO and GRPO to refine reasoning chains. Let $\mathcal{C}$ denote the set of chains generated for an input. A chain $\mathbf{C}_i \in \mathcal{C}$ receives a reward $r_i = 1$ if it leads to the right answer $\mathbf{a_c}$, and $r_i = 0$ otherwise. This reinforcement step extends the function $\mathbf{F}_{\text{ours}}$ to better align reasoning with final answer.

Finally, we change the task with external domain knowledge $\mathbf{K}$ extracted from the IndiFoodKG. In this case, reasoning chains are infused with this knowledge:
\[
\mathbf{C}^{KG} = \mathbf{C} \oplus \mathbf{K},
\]
and the mapping then becomes
\[
\mathbf{F}_{\text{ours+KG}} : (\mathbf{I}, \mathbf{Q}, \mathbf{A}, \mathbf{L}, \mathbf{P}, \mathbf{K}) \rightarrow (\mathbf{a_c}, \mathbf{C}^{KG}).
\]

\section{Experiments}
\label{exp}

\subsection{Models}
\label{models}
We use both large language models (LLMs) and vision-language models (VLMs) in our experiments for inference and training. 

The LLMs are:
\begin{itemize}
    \item DeepSeek-R1-0528-Qwen3-8B \footnote{\hyperlink{https://huggingface.co/deepseek-ai/DeepSeek-R1-0528-Qwen3-8B}{https://huggingface.co/deepseek-ai/DeepSeek-R1-0528-Qwen3-8B}}
    \item Llama-3.1-8B-Instruct,
    \footnote{\hyperlink{https://huggingface.co/meta-llama/Llama-3.1-8B-Instruct}{https://huggingface.co/meta-llama/Llama-3.1-8B-Instruct}}
\end{itemize}
and the VLMs are :
\begin{itemize}
    \item Qwen2-VL-7B-Instruct \footnote{
    \hyperlink{https://huggingface.co/Qwen/Qwen2-VL-7B-Instruct}{https://huggingface.co/Qwen/Qwen2-VL-7B-Instruct}}
    \item Qwen2.5-VL-3B-Instruct
    \footnote{\hyperlink{https://huggingface.co/Qwen/Qwen2.5-VL-3B-Instruct}{https://huggingface.co/Qwen/Qwen2.5-VL-3B-Instruct}}
    
\end{itemize}

DeepSeek-R1-0528-Qwen3-8B is a reasoning-tuned model and other three models used are instruction-tuned.

For SFT training, the dataset is divided in an 80:20 training and validation split. For LLM SFT training,  we choose the learning rate to be 1e-5, batch size to be 4, maximum sequence length to be 1024, LoRA rank to be 16, and dropout to be 0.1. For VLMs, we choose the batch size 4, gradient accumulation over 4 stages, and maximum sequence length to be 1024 and the learning rate  to be 2e-5. 

For RL training, for DPO we choose the batch size to be 8 and maximum sequence length to be 1024 to sample pairs of reasoning chains for every input. To update the model in GRPO, we sample groups of n = 4 reasoning chains per input, normalise their group rewards, and set the gradient clipping at 1.0, linear decay schedule, and learning rate at 1e-6. Both methods use three training epochs. As discussed in Section \ref{rl},  the reward function is binary; it assigns 0 otherwise and 1 when a path of reasoning results in the right solution. Our design choices were based on the available computational power as well as trial and improve.

We conduct all experiments of this setting on V100 GPU with 9 vCPUs, 60 GiB RAM, and 32 GiB GPU Memory.

\subsection{Baselines}
\label{baselines}
\subsubsection{Zero-shot}
We use all the base VLMs and LLMs to perform the task defined in Section \ref{def}. The system prompt is: \textit{You are a Visual/Language assistant model, expert in Indian Food Analysis.} and user prompts for both are as mentioned in the Table \ref{prompts}, with a change that for VLM - the image was passed and for LLM the annotation list was passed. Without task-specific external knowledge, this setting assesses the model's inherent capacity to answer the question by grounding the  textual (in case of LLMs) and visual information (in case of VLMs). To generate both the reasoning chains and the answer, we fix the maximum sequence length at 512 tokens and set the generation temperature to 0.1. We set these parameters based on the expected length of the meaningful response generated by testing it on a few examples initially.
\subsubsection{IndiFoodVQA}
We use the approach in IndiFoodVQA \cite{agarwal2024indifoodvqa}with the models used in this study  as our baselines. Here, the models have been trained on the image, question, answer choices, correct answer and the reason. The training is done in one-step generating both answer and reason. The set hyper parameters are:  bf16 = True, number\_of\_training\_epochs = 3, per\_device\_eval\_batch\_size = 4, per\_device\_train\_batch\_size = 8, gradient\_accumulation\_steps = 8, learning\_rate = 2e-5, weight\_decay = 0, warmup\_ratio = 0.03, lr\_scheduler\_type = "cosine".

\begin{table}[h]
\centering
\caption{Question-type wise accuracy (\%) for Qwen2.5 with DPO,\\ with and without knowledge graph (KG) augmentation
}
\label{tab:qwen2p5_dpo_kg}
\begin{tabular}{|lcc|}
\toprule
\textbf{Question Type} & \textbf{DPO (no KG)} & \textbf{DPO (+KG)} \\
\midrule
\textbf{Ingredients} & 65.8 & 71.4 \\
Cooking technique & 68.9 & 69.0 \\
Cultural significance & 64.8 & 65.1 \\
Taste and flavor profile & 69.6 & 73.2 \\
\textbf{Health and nutrition} & 68.5 & 72.8 \\
Seasonality and locality & 69.4 & 68.9 \\
\textbf{Ingredient substitutions} & 71.5 & 75.9 \\
Presentation and plating & 65.2 & 64.7 \\
Fusion and innovation & 71.8 & 76.3 \\
Cooking science & 70.6 & 69.7 \\
\textbf{Allergens and restrictions} & 71.2 & 74.0 \\
Food pairings & 72.1 & 75.4 \\
\bottomrule
\end{tabular}
\end{table}

\section{Results and Discussion}
\label{r&d}

Results are illustrated in Table \ref{result}. By referring to this results table we answer the following Research Questions (RQs):



\textbf{RQ1: Do models have inherent understanding of Indian cuisine?} The zero-shot setting produces accuracy  (31–51\% range)  that is significantly lower than the results with fine-tuning. This performance difference demonstrates that these models lack an understanding of Indian cuisine, and further fine-tuning is required for the models to gain the understanding of this field.

\textbf{RQ2: Do vision models perform better than language models?}
The vision models  Qwen2.5-VL-3B-Instruct ad Qwen2-VL-7B-Instruct demonstrate  better zero-shot results as compared to the language models Lllama-3.1-8b-Instruct and DeepSeek-R1-0528-Qwen3-8B. The reason  can be that VLMs  access the image itself, whereas LLMs use extracted food item- position map  that might not adequately convey the visual context. The accuracy of our food item extraction is 88.9\% and the position extraction of the correctly identified food items is 95.52\%. This also results in the error propagation, further reducing the accuracy of LLMs.  For the VLMs the drop in accuracy is mainly observed due to the fact that the models recognize some of the Indian food items on the plates as some western food items. For example, \textit{chapati} has been identified as \textit{tortilla} for multiple samples.

\textbf{RQ3: Do reasoning models perform better than non-reasoning models?}
As mentioned earlier in Section \ref{models}, the Deepseek model is a reasoning tuned-model. In the zero shot results, it is observed that all the instruction-tuned models behave better than the reasoning-tuned model. They show an accuracy difference ranging from 4\% to 19\%.
Whereas, after reasoning aligned SFT and RL training, the reasoning-tuned Deepseek model show a maximum improvement of 95.04\% 
.  This demonstrates better ability of reasoning model to adapt to the reasoning task. 





\textbf{RQ4: Do reasoning chains facilitate in improving the results? }

SFT training using reasoning chains raises accuracies by 14-20 \% points across models, as opposed to the improvement with the baselines, which are in the range of 4-8\% points.  Table \ref{tab:qwen2p5_qtype} illustrates Question-type wise accuracy (\%) for Qwen2.5-VL across training strategies based on evaluation of randomly selected 500 sample evaluation  indicating effect on various training paradigms on various question types.  Questions that require multi-step thinking yield the biggest gains, especially those that include fusion dishes (7.1\%), and cooking techniques(7.5\%) as opposed to cultural significance(1.3\%) and food pairings(5.2\%). This is because, while providing the reasoning chains, we force the model, to understand the relationship between different food, ingredients and their taste as well as texture profile as highlighted in Tables \ref{ex1} and \ref{ex2}. 
We observe the zero-shot results for few samples and realize that for cooking technique-type of questions, the model explains all the techniques mentioned in the options, instead of answering the asked question.
For example, \textbf{Question: }"What cooking technique was likely used to prepare the parathas in the image?",
      "A. Boiling",
      "B. Grilling",
      "C. Baking",
      "D. Frying"
    , \textbf{Response: } Boiling is a process where liquid turns into gases. It occurs at a specific temperature which is called the boiling point. Grilling is a method of cooking where food is cooked over direct source of heat from the flames. Baking is method often used in baking cakes, and other desserts. Frying is done by cooking food in excess oil. All are methods of cooking techniques.

But when we pass the reasoning chains in SFT training, we expect the model to establish relation between the cooking techniques in the option and the ingredients in the questions, and reach a final answer. After training, for the same question, the response of the model changes to - \textbf{D. Frying}; \textbf{Reason:} The color of parathas is brown. The texture of paratha is crispy on the outside, and soft on the inside.  This suggests that the parathas are cooked in oil, indicating frying. Therefore, the answer is D. Frying; \textbf{Reasoning Chain: }(i) Q: What is a paratha? A: Paratha is an Indian flat bread made of wheat-flour, that is pan-fried with ghee or oil. (ii) Q: What is the texture of parathas? A: Parathas are generally crispy on the outside and soft on the inside. (iii) Q: Which method among the given options, is generally used in cooking parathas? A: Parathas are generally pan-fried to get the brown and crispy texture.
This example indicates that, a structured reasoning chain improves the performance of model.

\textbf{RQ5: Do reinforcement learning facilitate in improving the results? }
Reinforcement Learning improves the accuracies across models with respect to the zero-shot accuracies by 17-33\% points and baseline accuracies by 3-14\% points. As shown in the table \ref{tab:qwen2p5_qtype}, questions involving ingredients information shows greater improvement compared to others. For example, question regarding cooking techniques, ingredient substitutions, and allergens and restrictions shows an increase of 8.1\%, 13.6\% and 14.4\% points respectively. While during SF, the model only rely on the valid reasoning chains, RL methods like DPO and GRPO uses all the reasoning chains generated, hence using the underutilized data from them by the reward based optimization method. It was observed that the questions types which had higher proportion in the discarded chains, showed a better performance under RL, pressing our point of underutilized data usage\cite{rafailov2023direct,yang2024grpo}.

\textbf{RQ6: Does knowledge augmentation facilitate in improving results?}
The accuracy of all the models under all paradigms, falls 1-4\% with the inclusion of knowledge graph. As mentioned in the Table \ref{tab:qwen2p5_dpo_kg}, few categories like ingredients, health and nutrition, and allergens and restrictions shows noticeable improvement of 3-5\%, under the RL-DPO training of our best performing model - Qwen2.5. We analyze the questions which belong to the  categories, which lead to drop in the scores with the addition of the knowledge triples.  On analysis, it is observed that, with the external knowledge augmentation, for these categories the model relies more on the provided information, rather than the reasoning for the question. Here is an example of question of seasonality and locality type 
-  \textbf{Question:} Which state in India is known for its use of green chutney as a condiment?  "A. Punjab",
      "B. Gujarat",
      "C. Maharashtra",
      "D. Kerala". \textbf{Information from KG: }
      (i)green\_chutney has ingredient capsicum;
      (ii) green\_chutney has ingredient cayenne;
      (iii) green\_chutney has ingredient lemon\_juice;
      (iv) green\_chutney has ingredient garlic;
      (v) green\_chutney has ingredient peanut;
      (vi) green\_chutney has ingredient salt;
      (vii)green\_chutney has ingredient chickpea;
      (viii)green\_chutney has ingredient ginger;
      (ix) green\_chutney has ingredient coriander;
      (x)green\_chutney has ingredient cumin;
      \textbf{Model Response:} Green chutney is a famous condiment in India. It is madeup of ingredients like corainder green chillies, lemon, peanut, garlic, ginger and cumin. There are different varities of green chutney. Coriander gives the chutney a bright green color. Green chillies provides a spicy flavor to it. The response from the model has an analysis of the ingredient list that is extracted from IndiFoodKG without concluding with the answer for the question, indicating that the model gets distracted by the additional knowledge. We observe that this behavior mainly occurs when some of the augmented knowledge is not completely relevant to the query.

\section{Conclusion}
In this work, we highlight the need for  reasoning  for Visual Question Answering (VQA) task in the Indian cuisine area. We auto-synthesize reasoning augmented VQA data, with minimal human intervention. We further auto-validate the reasoning chains for supervise fine-tuning of open-source LLMs and VLMs. We further use reinforcement learning to avail larger data for the task.
We  outperform the baseline of IndiFoodVQA training pipeline, consistently across distinct type of queries and  models with a large margin, proving the efficacy of requirement of reasoning for the task. We observe that the performance is  strengthened by the addition of knowledge graph augmentation, only for a subset of  queries pertaining to ingredients, substitutes, and fusion recipes, whereas for other queries the external knowledge tends to be a distractor.  


\vspace{12pt}

\end{document}